\documentclass[conference]{IEEEtran}
\IEEEoverridecommandlockouts

\usepackage{cite}
\usepackage{amsmath,amssymb,amsfonts}
\usepackage{algorithmic}
\usepackage{graphicx}
\usepackage{textcomp}
\usepackage{xcolor}
\usepackage{multirow}
\usepackage{jabbrv}

\def\BibTeX{{\rm B\kern-.05em{\sc i\kern-.025em b}\kern-.08em
    T\kern-.1667em\lower.7ex\hbox{E}\kern-.125emX}}

\makeatletter 
\newcommand{\linebreakand}{%
\end{@IEEEauthorhalign}
\hfill\mbox{}\par
\mbox{}\hfill\begin{@IEEEauthorhalign}
}
\makeatother 

\begin{document}

\title{Toward Practical BCI: A Real-time Wireless Imagined Speech EEG Decoding System\\
\thanks{This work was the result of project supported by Korea University - KT (Korea Telecom) R\&D Center, and partially supported the Institute of Information \& Communications Technology Planning \& Evaluation (IITP) grant, funded by the Korea government (MSIT) (No. RS-2019-II190079, Artificial Intelligence Graduate School Program (Korea University))}
}

\author{
\IEEEauthorblockN{Ji-Ha Park}
\IEEEauthorblockA{\textit{Dept. of Artificial Intelligence} \\
\textit{Korea University} \\
Seoul, Republic of Korea \\
jiha\_park@korea.ac.kr}
\and
\IEEEauthorblockN{Heon-Gyu Kwak}
\IEEEauthorblockA{\textit{Dept. of Artificial Intelligence} \\
\textit{Korea University} \\
Seoul, Republic of Korea \\
hg\_kwak@korea.ac.kr}
\and
\IEEEauthorblockN{Gi-Hwan Shin}
\IEEEauthorblockA{\textit{Dept. of Brain and Cognitive Engineering} \\
\textit{Korea University} \\
Seoul, Republic of Korea \\
gh\_shin@korea.ac.kr}
\and
\IEEEauthorblockN{Yoo-In Jeon}
\IEEEauthorblockA{\textit{Tech. Innovation Group} \\
\textit{KT R\&D Center} \\
Seoul, Republic of Korea \\
yooin.jeon@kt.com}
\and
\IEEEauthorblockN{Sun-Min Park}
\IEEEauthorblockA{\textit{Tech. Innovation Group} \\
\textit{KT R\&D Center} \\
Seoul, Republic of Korea \\
sunmin.park@kt.com}
\and
\IEEEauthorblockN{Ji-Yeon Hwang}
\IEEEauthorblockA{\textit{Tech. Innovation Group} \\
\textit{KT R\&D Center} \\
Seoul, Republic of Korea \\
jiyeon.hwang@kt.com}
\and
\IEEEauthorblockN{Seong-Whan Lee}
\IEEEauthorblockA{\textit{Dept. of Artificial Intelligence} \\
\textit{Korea University} \\
Seoul, Republic of Korea \\
sw.lee@korea.ac.kr}
}

\maketitle

\begin{abstract}
Brain-computer interface (BCI) research, while promising, has largely been confined to static and fixed environments, limiting real-world applicability. To move towards practical BCI, we introduce a real-time wireless imagined speech electroencephalogram (EEG) decoding system designed for flexibility and everyday use. Our framework focuses on practicality, demonstrating extensibility beyond wired EEG devices to portable, wireless hardware. A user identification module recognizes the operator and provides a personalized, user-specific service.
To achieve seamless, real-time operation, we utilize the lab streaming layer to manage the continuous streaming of live EEG signals to the personalized decoder. This end-to-end pipeline enables a functional real-time application capable of classifying user commands from imagined speech EEG signals, achieving an overall 4-class accuracy of 62.00~\% on a wired device and 46.67~\% on a portable wireless headset. This paper demonstrates a significant step towards truly practical and accessible BCI technology, establishing a clear direction for future research in robust, practical, and personalized neural interfaces.
\end{abstract}


\section{INTRODUCTION}
Brain-computer interfaces (BCIs) based on non-invasive neural signals such as electroencephalogram (EEG) have garnered significant attention as a technology for decoding human intent directly from neural activity~\cite{kim2015abstract, peksa2023state}. Much of the research in this field has focused on translating these complex neural patterns into actionable commands for external devices~\cite{bulthoff2003biologically, lee2018deep}. Among the various approaches, endogenous paradigms, which utilize internally generated brain signals like imagined speech, offer a highly intuitive and natural means of communication without reliance on external stimuli~\cite{willett2023high, park2025towards}. These systems, which aim to translate a user's thoughts into commands, hold immense promise for assistive technologies and novel interaction methods~\cite{lee2003pattern}. However, a notable gap remains between laboratory demonstrations and their deployment as practical, everyday applications~\cite{metzger2023high, prabhakar2020framework}.

The primary challenge hindering the real-world deployment of BCI systems is their limited practicality~\cite{saha2021progress}. One of the main issues is the extensive, user-specific data required for calibration, a process that is not only time-consuming but also creates a barrier for new users. Moreover, research has predominantly been conducted in restricted settings using high-density, wired EEG device, which is impractical for daily life and limits scalability~\cite{ding2013changes}. A gap persists between high-performance models developed on stationary, wired equipment and the demands of practical, wireless hardware~\cite{suk2014predicting, kaifosh2025generic}. To bridge this gap, it is crucial to develop unified systems that are not only validated for high-fidelity performance on wired devices but are also proven to be lightweight and robust enough for real-time, practical use on portable wireless headsetst~\cite{lee1996multiresolution, lee2015motion, eldawlatly2024role}.

In this paper, we propose a real-time wireless imagined speech EEG decoding system designed towards practical BCI. Our proposed method performs user identification to recognize the operator, which in turn loads a user-specific pre-trained model, and then proceeds to intention classification based on that personalized model~\cite{kwak2025towards}. This unified framework is designed to bridge the gap, operating effectively on both high-fidelity wired setups and practical wireless headsets.
To manage the continuous data flow required for real-time inference, the system utilizes the lab streaming layer (LSL)~\cite{kothe2025lab, cho2021neurograsp}. Emphasizing practicality, the system is designed for a sparse minimal channel configuration and is validated on both high-fidelity wired equipment and a portable, wireless device~\cite{ishtiaque2025systematic}. This work assesses the system's potential for transitioning from controlled settings to everyday life, paving the way for more accessible and practical BCI applications.

\begin{figure*}[t]
\centerline{\includegraphics[width=0.99\textwidth]{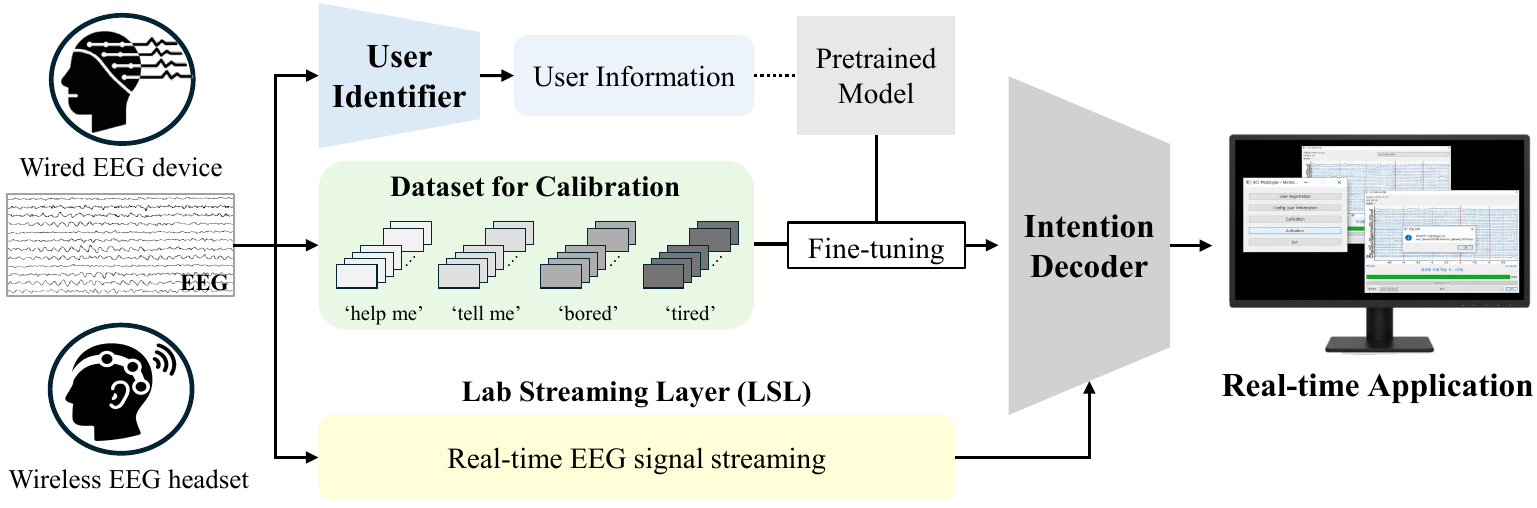}}
\caption{Overall framework of a real-time practical BCI system. Imagined speech EEG signals are acquired from either a wired (Brain Products) or wireless (Emotiv EPOC X) device. The user identifier module first processes the signals to recognize the user's information. Based on this identity, a corresponding pre-trained model is loaded and subsequently fine-tuned using a small set of newly acquired calibration data. This fine-tuned model then serves as the intention decoder. For real-time operation, live EEG signals are streamed via LSL directly to the personalized intention decoder, which performs inference to classify the user's intention for the final application display.}
\label{fig1}
\end{figure*}

\section{MATERIALS AND METHODS}
\subsection{Data Acquisition}
\subsubsection{Experimental setup}
Three healthy Korean participants in this study. The entire experimental procedure received ethical approval from the Institutional Review Board of Korea University [KUIRB–2024–0065–01] after a thorough review. The experiment was focused on the speech imagery paradigm, an endogenous EEG task. The participant was instructed to internally articulate four command-oriented words, chosen for their relevance to practical BCI applications (C1: `help me', C2: `tell me', C3: `bored', and C4: `tired'). The session consisted of 800 trials in total, with 200 trials recorded for each of the classes. Each trial had a duration of two seconds, during which the participant performed the imagery task.

\subsubsection{EEG recording and preprocessing}
As shown in Fig.~\ref{fig1}, for the primary data collection, we utilized a high-precision setup~\cite{kwak2025towards}. EEG signals were acquired using a BrainAmp amplifier (Brain Products GmbH, Germany). A total of 32 channels were used in accordance with the International 10/20 system for electrode placement. Out of these, 30 EEG channels were affixed to the scalp using conductive gel, and two channels (`TP9' and `TP10'), were used as electrooculogram (EOG) channels. The `Fpz' channel served as the ground, while the reference electrode was positioned on the nasal bridge. The data was recorded via BrainVision software at a 250~Hz sampling rate. For preprocessing, we applied a 60~Hz notch filter to attenuate power-line interference, and no additional filtering was performed~\cite{suk2011subject}. Subsequently, because the amplitude of raw EEG is on the microvolt ($\mu V$) scale, which can make the model excessively sensitive to small weight and bias updates, we rescaled the signal amplitudes by $10^4$ to stabilize optimization and improve numerical robustness.

To further assess the practical viability of our framework in real-world settings, we conducted an additional validation using the Emotiv EPOC X, a consumer-grade portable EEG device~\cite{tarara2025motor}. This 12-channel wireless headset, utilizing the `F7', `F3', `FC5', `T7', `P7', `O1', `O2', `P8', `T8', `FC6', `F4', and `F8' electrodes, is designed for mobile applications and represents a more realistic hardware scenario for everyday BCI use. The data collected from this device allowed us to evaluate our system's performance and robustness outside the controlled laboratory environment.

\begin{figure*}[t]
\centerline{\includegraphics[width=0.97\textwidth]{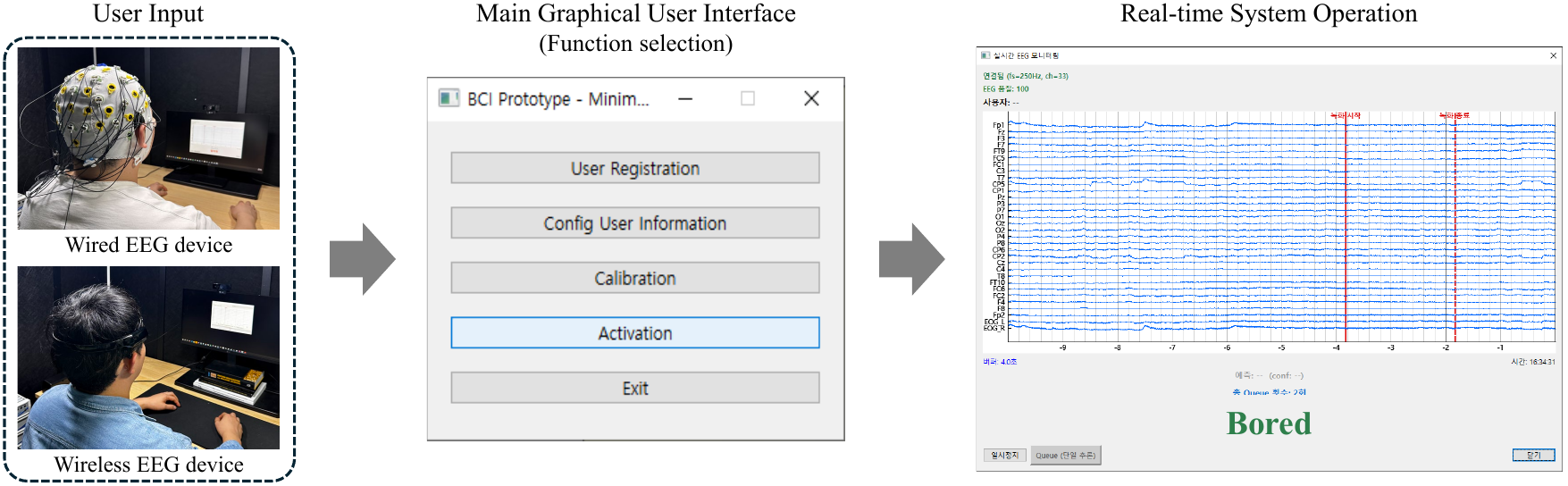}}
\caption{Overview of the proposed BCI application workflow. Users can select either a wired or wireless device for EEG signal input. The main GUI is used to select functions such as user registration, calibration, or system activation. The system is designed to visualize EEG signals in real-time and display the corresponding inference results.}
\label{fig2}
\end{figure*}

\subsection{Decoding Model}
The EEG decoding model is designed based on a dual sequential transformer encoder architecture to effectively learn the spatio-temporal properties inherent in EEG signals~\cite{lee1995multilayer, lee1999integrated, jiang2025decoding}. The incoming multi-channel EEG and EOG signals are passed through a 1D-convolutional block. This patchification step serves to extract local temporal features from each channel's signal, transforming the raw signals into a series of feature vector sequences that the transformer can process.

These generated patch sequences are first fed into a temporal transformer encoder~\cite{lee1997new}. This encoder models the complex dynamic relationships between patches along the time axis within each channel. Through this process, a compressed feature vector of each channel's temporal information is created. These feature vectors are passed to a spatial transformer encoder. This second encoder learns the complex spatial inter-dependencies between different channels. Furthermore, the model encodes relative positional information within its attention mechanism, allowing it to learn the relationships of time and channels more flexibly than with fixed positional embeddings. Finally, this representation vector, which integrates all spatio-temporal features, is passed to a classifier to be decoded into the user-specific features for user identification and features to classify the user's imagined speech intention~\cite{jiang2025decoding}.

\subsection{Personalized BCI Application}
\subsubsection{User identification}
User identification task is the process of ascertaining the identity of the current user operating the application~\cite{sun2019eeg, lee2020continuous, kwak2025towards}. This task serves as a critical prerequisite for personalization, as it allows the system to access and apply user-specific preferences and settings. Our approach leverages the inherent subject variability characteristic of EEG signals. This phenomenon refers to the distinctiveness of neural patterns across different individuals, even when they are in the same environment. By exploiting this inter-subject variability, our system extracts unique neural signatures that function as biomarkers for recognizing the user, thereby enabling the BCI to adapt its services to the specific individual.

\subsubsection{Intention classification}
Intention classification task is focused on decoding the user's desired action or command directly from their brain activity~\cite{kwak2025towards}. As a direct reflection of the brain's cognitive processes, EEG signals inherently contain rich information about an individual's mental state and intent. To facilitate a natural and self-directed mode of communication, our system uses endogenous EEG paradigms, such as speech imagery. In contrast to exogenous paradigms, which depend on external sensory stimuli, endogenous paradigms are generated solely by the user's imagination, thus offering a more intuitive method for interacting with the BCI application.

\subsubsection{System implementation}
As shown in Fig.~\ref{fig2}, to validate the practical utility of our proposed framework, we developed a real-time BCI application that provides an interactive experience for the user~\cite{vidyaratne2017real}. The application's workflow is designed for flexible application and user-specific adaptation. Initially, the system initiates a brief setup phase through the main graphical user interface (GUI). It begins by recording 30 seconds of the user’s stable, resting-state EEG signals, which are used to establish a unique neural signature for subsequent user identification. Following this, the system enters a calibration mode, where it collects a minimal dataset consisting of only ten trials per class of imagined speech. This small dataset is immediately used to fine-tune the pre-trained model, adapting it to the specific user's neural patterns and enabling immediate, personalized use of the BCI~\cite{wu2020transfer}.

Once the setup is complete, the application transitions to its real-time inference mode. It continuously acquires EEG data from the user in 0.1-second intervals, accumulating these segments into a data buffer. This process includes a real-time visualization module, allowing the user to monitor their own brainwave activity. For command decoding, two-second sliding window is applied to the buffer, ensuring that each classification is based on a sufficient temporal segment of data. The system performs inference on this window to classify the user's intention. If a specific command other than the `rest' state is detected, the classification result is immediately displayed on the user interface, providing direct feedback and enabling real-time control.

\begin{table}[t]
\setlength{\tabcolsep}{3.1pt}
\renewcommand{\arraystretch}{1.2}
\caption{Intention classification performance on imagined speech EEG.}
\begin{center}
\begin{tabular}{cccccc}
\hline
\textbf{Device}                    & \textbf{Class}         & \textbf{Precision $\uparrow$} & \textbf{Recall $\uparrow$}    & \textbf{F1-score $\uparrow$}  & \textbf{Acc. (\%) $\uparrow$} \\ \hline
\multirow{5}{*}{\textbf{Wired}}    & C1             & 0.65 $\pm$ 0.15          & 0.73 $\pm$ 0.09          & 0.68 $\pm$ 0.11          & -                        \\
                                   & C2             & 0.63 $\pm$ 0.19          & 0.67 $\pm$ 0.13          & 0.64 $\pm$ 0.16          & -                        \\
                                   & C3             & 0.53 $\pm$ 0.08          & 0.43 $\pm$ 0.14          & 0.47 $\pm$ 0.12          & -                        \\
                                   & C4             & 0.71 $\pm$ 0.07          & 0.65 $\pm$ 0.15          & 0.67 $\pm$ 0.09          & -                        \\
                                   & \textbf{Avg.}  & \textbf{0.63 $\pm$ 0.05} & \textbf{0.62 $\pm$ 0.03} & \textbf{0.62 $\pm$ 0.04} & \textbf{62.00 $\pm$ 3.47}    \\ \hline
\multirow{5}{*}{\textbf{Wireless}} & C1             & 0.39 $\pm$ 0.12          & 0.51 $\pm$ 0.22          & 0.44 $\pm$ 0.16          & -                        \\
                                   & C2             & 0.57 $\pm$ 0.13          & 0.43 $\pm$ 0.26          & 0.43 $\pm$ 0.14          & -                        \\
                                   & C3             & 0.45 $\pm$ 0.16          & 0.49 $\pm$ 0.19          & 0.44 $\pm$ 0.10          & -                        \\
                                   & C4             & 0.62 $\pm$ 0.05          & 0.45 $\pm$ 0.07          & 0.51 $\pm$ 0.04          & -                        \\
                                   & \textbf{Avg.}  & \textbf{0.51 $\pm$ 0.04} & \textbf{0.47 $\pm$ 0.00} & \textbf{0.46 $\pm$ 0.01} & \textbf{46.67 $\pm$ 0.72}    \\ \hline
\end{tabular}
\label{tab1}
\end{center}
\footnotesize{$^*$Acc.: accuracy, $^*$Avg.: average}
\end{table}

\section{RESULTS AND DISCUSSION}
\subsection{Performance on Wired EEG Device}
Table~\ref{tab1} displays the quantitative performance of intention decoding from imagined speech EEG. In the controlled laboratory setting using the wired device (Brain Products), our system demonstrated significant and meaningful performance. The model achieved an overall classification accuracy of $62.00 \pm 3.47~\%$ in the four classes, which is above the theoretical chance level (25~\%). The macro-average F1-score was $0.62 \pm 0.04$, indicating a robust and balanced classification capability across all commands.
These results validate the fundamental architecture of our spatio-temporal decoding model and confirm the feasibility of our proposed pipeline, including the user identification and rapid fine-tuning process, in a high-fidelity environment.

\subsection{Performance on Wireless EEG Device}
When deploying the same framework on the portable wireless device (Emotiv EPOC X), the system achieved an overall accuracy of $46.67 \pm 0.72~\%$ and a macro-average F1-score of $0.46 \pm 0.01$. These metrics are lower than those of wired lab-grade equipment, which can be attributed to the inherent challenges of consumer-grade hardware, such as lower signal-to-noise ratio and potential signal instability. However, this performance remains above chance level. This result is highly encouraging, as it demonstrates the potential of our system to function outside of a controlled lab. It confirms that our lightweight, 12-channel model architecture is viable for practical, real-world scenarios and establishes a solid baseline for future improvements in performance and robustness on portable hardware.

\section{CONCLUSIONS}
In this paper, we introduced a practical, real-time BCI system designed to enhance user convenience and provide personalized experiences, demonstrating its potential for broad, general-purpose applications. Using a real-time EEG signal streaming method, our system performs simultaneous user identification and 4-class classification from imagined speech EEG. Our experiments confirmed the system's high efficacy on a wired device and, more importantly, demonstrated its viability on a portable headset, thereby bridging the critical gap between laboratory research and real-world applicability. While the performance on the portable device indicates a promising future, it also highlights the need for further refinement. Future work will therefore focus on enhancing the decoding accuracy and confidence of our system on such portable hardware, a crucial step towards realizing truly accessible and reliable BCI technology for practical use.

\bibliographystyle{jabbrv_IEEEtran}
\bibliography{REFERENCE_jh}

\end{document}